\documentclass{article}

\usepackage{arxiv}

\usepackage[utf8]{inputenc} 
\usepackage[T1]{fontenc}    
\usepackage{hyperref}       
\usepackage{url}            
\usepackage{booktabs}       
\usepackage{amsfonts}       
\usepackage{nicefrac}       
\usepackage{microtype}      
\usepackage{lipsum}		
\usepackage{graphicx}
\usepackage{natbib}
\usepackage{doi}
\usepackage{subcaption}
\usepackage{graphicx}
\usepackage{float}

\title{Lesion Detection on Leaves using Class Activation Maps}

\author{ \href{https://orcid.org/0009-0003-0094-2292}{\includegraphics[scale=0.06]{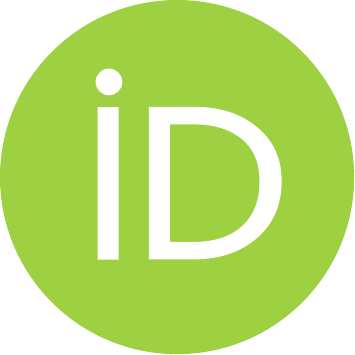}\hspace{1mm}Enes Sadi Uysal}\\
	Huawei R\&D Turkiye\\
	\texttt{enes.sadi.uysal1@huawei.com} \\
	\And
	\href{https://orcid.org/0000-0002-5175-3438}{\includegraphics[scale=0.06]{orcid.pdf}\hspace{1mm}Deniz Sen} \\
	Huawei R\&D Turkiye\\ 
	\texttt{deniz.sen1@huawei.com} \\
	\And
	\href{https://orcid.org/0000-0001-7254-9316}{\includegraphics[scale=0.06]{orcid.pdf}\hspace{1mm}Ahmet Haydar Ornek}\\
	Huawei R\&D Turkiye\\
	\texttt{ahmet.haydar.ornek2@huawei.com} \\
    \And
	\href{https://orcid.org/0000-0001-9230-1617}{\includegraphics[scale=0.06]{orcid.pdf}\hspace{1mm}Ahmet Emin Yetkin}\\
	Huawei R\&D Turkiye\\
	\texttt{ahmet.emin.yetkin@huawei.com} \\
}

\hypersetup{
pdftitle={Lesion Detection on Leaves using Class Activation Maps},
pdfsubject={},
pdfauthor={Enes S. Uysal, Deniz Sen, Ahmet H. Ornek, Ahmet E. Yetkin},
pdfkeywords={Class Activation Maps,  Lesion Detection, CNN},
}

\begin{document}
\maketitle

\begin{abstract}
	Lesion detection on plant leaves is a critical task in plant pathology and agricultural research. Identifying lesions enables assessing the severity of plant diseases and making informed decisions regarding disease control measures and treatment strategies. To detect lesions, there are studies that propose well-known object detectors. However, training object detectors to detect small objects such as lesions can be problematic. In this study, we propose a method for lesion detection on plant leaves utilizing class activation maps generated by a ResNet-18 classifier.
    In the test set, we achieved a 0.45 success rate in predicting the locations of lesions in leaves. Our study presents a novel approach for lesion detection on plant leaves by utilizing CAMs generated by a ResNet classifier while eliminating the need for a lesion annotation process.

\end{abstract}

\keywords{Class Activation Maps \and Lesion Detection \and CNN}

\section{Introduction}
Lesion detection on plant leaves is a critical task in plant pathology and agricultural research. Plant diseases, caused by various pathogens such as fungi, bacteria, and viruses, can significantly impact crop yield, quality, and overall agricultural productivity. Early and accurate detection of lesions on plant leaves plays a crucial role in disease management and prevention, allowing farmers and researchers to implement timely interventions and minimize the spread of infections.

Lesions are visible abnormalities or discolorations on plant leaves, which can manifest as spots, blotches, necrotic areas, or other irregular patterns. They are often indicators of disease presence or environmental stressors, such as nutrient deficiencies, chemical damage, or physical injuries. Identifying and analyzing these lesions enable scientists and growers to assess the severity of plant diseases, monitor their progression, and make informed decisions regarding disease control measures and treatment strategies.

Traditionally, lesion detection has relied on manual visual inspection by trained experts, which can be time-consuming, labor-intensive, and prone to subjective interpretation. However, recent advancements in imaging technologies, machine learning algorithms, and computer vision techniques have opened up new possibilities for automating and enhancing lesion detection processes. By leveraging these technological tools, researchers are able to develop efficient and accurate methods for lesion identification and classification, aiding in the early diagnosis and management of plant diseases.

ResNet \cite{he2015deep}, short for Residual Neural Network, is a deep learning architecture known for its exceptional performance in image classification tasks. It incorporates skip connections, or residual connections, which help alleviate the vanishing gradient problem and enable the network to learn more effectively. By using a pre-trained ResNet model as a feature extractor, we can leverage the knowledge learned from vast amounts of labeled data, enhancing the overall performance and robustness of the lesion detection system. In this study, we propose a method for lesion detection on plant leaves utilizing class activation maps generated by a ResNet-18 classifier. Class activation maps (CAMs) \cite{zhou2016learning} are powerful visualizations that highlight the regions in an image that contribute most to a particular class prediction made by a convolutional neural network (CNN). By leveraging CAMs in combination with the ResNet-18 architecture, we aim to accurately identify and localize lesions on plant leaves, providing a cost-efficient tool for plant pathology and agricultural research. We reduce the complexity of the problem from detection to classification considering both training and computational costs of classifiers are lower than object detector models. In addition, we get rid of drawing separate bounding boxes for each lesion. 

\section{Literature Review}
There have been many approaches to plant disease detection; specifically, image processing, machine learning, and deep learning algorithms were highly investigated to solve the related problems. Segmentation via color channel thresholding is a basic, yet effective technique in image processing, and \cite{dhaygude2013agricultural} utilized it to retrieve the most relevant segments, which are then put through statistical analysis. Similarly, \cite{ngugi2021new} created a pipeline that involves Otsu segmentation \cite{otsu} on 3 different color spaces and an empirically adapted rule mechanism. Apart from the traditional image segmentation techniques, \cite{genetic} used the genetic algorithm to separate the diseased parts of the leaf.

Nowadays as the data got denser, deep feature extraction is preferred over traditional methods due to their complexity and generalization capability. \cite{korkut2018detection} compared the performances of different classification algorithms which were trained over the deep features of a pretrained InceptionV3 feature extractor. As the amount of data increases, imbalance problems between the classes can occur, which is a problem \cite{9568965} addressed by data augmentation and the introduced stepwise transfer learning scheme. \cite{ashwinkumar2022automated} proposed a multistage deep classifier that took advantage of the efficiency of MobileNet in the feature extraction step. \cite{wang2021t} introduced a trilinear CNN model architecture that separates the crop and disease identification processes which leads the network to learn more distinct and informative features. \cite{atila2021plant} compared the state-of-the-art deep classifier performances over the standard and augmented PlantVillage dataset \cite{DBLP:journals/corr/HughesS15}. Following the developments in the singular model configurations, \cite{agriculture11070617} adapted an ensemble model scheme which is constituted from DenseNet121 \cite{huang2017densely}, EfficientNetB7 and EfficientNet Noisy Student models \cite{tan2020efficientnet}. Furthermore, \cite{li2022improved} trained a YOLOV5s-based object detector to localize and classify multiple diseased plants inside a scene. 

Besides supervised object detectors such as YOLO, classification models can be used to detect objects by utilizing activation maps and this concept is known as Weakly Supervised Object Detection \cite{wsod1}. When a classification model is trained, its last convolutional layers learn high-level features related to desired classes. CAM is a technique that highlights regions of desired classes in images using the last convolutional layers. \cite{gradcamplusplus1} uses a combination of first-order and second-order gradients to calculate the importance weights of each feature map. This allows for more accurate localization of desired classes in images. \cite{gradcam1} combines selective search and gradient-based CAM (GradCAM) to achieve a higher object detection performance using IoU. When PASCAL VOC 2007 and 2012 dataset and MS COCO dataset are used, their method significantly improves their IoU. \cite{wsod_seg} proposes a generative adversarial segmentation module to obtain more precise localization including segmentation tasks. 

\section{Materials and Method}

\subsection{Dataset}
The final dataset consists of the alteration and combination of several public datasets. PlantVillage is an image classification dataset containing 54304 samples and 38 classes; the classes are crops taken from 14 species, each having images from healthy and diseased lab-controlled samples. The dataset severely suffers from class imbalance, such that the class with the most samples is constituted of 5507 images, whereas the one with the least number of samples includes 152 images.

The second dataset we used in this work is the Leaf Disease Segmentation Dataset \cite{Alam_2021} which includes 588 images of diseased leaves and 588 binary masks segmenting the diseased parts of the corresponding leaf. The samples are also augmented using basic transformations such as rotation and flipping. Using this dataset, we created an object detection dataset where each lesion segment is replaced with its smallest enclosing rectangle, which became the ground truth bounding box. 

Finally, as we train both image classifiers and object detectors for plant lesions, we combined our 2 datasets to obtain a new image classification dataset; we counted each image of the newly obtained lesion detection dataset as a positive sample, on the other hand, we took the healthy plant images from the PlantVillage dataset and counted each one as a negative sample, as they do not include any lesions.

\subsection{Class Activation Mapping (CAM)}

To explain the classification process of a CNN model, visual explanation techniques such as CAM are used. The CAM techniques aim to emphasize significant areas of provided images by utilizing the final convolutional layers of the models, which include high-level features. The initial CAM technique needs to be adjusted by eliminating the fully-connected layer, incorporating the Global Average Pooling layer, and retraining the model. Once this is done, the activation map related to training can be generated using Eq. \ref{eq:cam}.

\begin{equation}
   L_{}^{c}=\sum_{}w_{}^{c}A
   \label{eq:cam}
\end{equation}

Where $L_{}^{c}$ represents the activation map for class $c$, $w_{}^{c}$ represents the importance of the convolutional layer for class $c$, and $A^{}$ represents the convolutional layer. As the model's performance decreases when the fully-connected layer is eliminated,  GradCAM techniques have been introduced. GradCAM eliminates the need for layer removal and retraining. To determine the weights for GradCAM, Eq. (\ref{eq:gradcam}) is used. Once the weights are calculated, the activation maps are generated through the Eq \ref{eq:cam}.

\begin{equation}
    w_{}^{c}=\frac{1}{N}\sum_{}\frac{\partial Y^{c}}{\partial A_{}^{}}
   \label{eq:gradcam}
\end{equation}

\subsection{Model Architecture and Training}
We have chosen ResNet-18 as the diseased plant classifier, thanks to its relatively low parameter count, and proven learning capability. 

The ResNet-18 \cite{he2015deep} architecture consists of 18 layers, including a convolutional layer, a max pooling layer, and several residual blocks. Each residual block contains two convolutional layers and a shortcut connection that bypasses the convolutional layers. The shortcut connection allows the gradient to flow directly through the network, which helps to prevent the vanishing gradient problem that can occur in very deep networks.

The first layer of the network is a 7x7 convolutional layer with stride 2, followed by a 3x3 max pooling layer with stride 2. This reduces the spatial dimensions of the input by a factor of 4. The next 4 layers each contain a single residual block, with 2 convolutional layers and a shortcut connection. The final layer of the network is a fully connected layer with 1000 output units, which corresponds to the number of classes in the ImageNet dataset.

The ResNet-18 architecture has been shown to achieve state-of-the-art performance on a variety of image classification tasks, including the ImageNet dataset \cite{imagenet}. It is also relatively lightweight compared to other deep neural network architectures, making it well-suited for deployment on resource-constrained devices.

We trained the model for 4 epochs on the final lesion dataset with a batch size of 128 and input size of 224x224. We used Adam optimizer with an initial learning rate of 0.001, $\beta_1$, and $\beta_2$ parameters of 0.9 and 0.999 respectively.

\subsection{Extracting The Bounding Boxes from Activation Map}
The present study utilizes the Class Activation Mapping (CAM) technique to extract the activation map. In order to segment the highly attended regions, a combination of OTSU and binary thresholding methods is employed, resulting in the creation of binary maps. Subsequently, an opening operation is applied with a 3 by 3 kernel and 3 iterations to separate the lesions from each other. Finally, the contours are detected using the chain approximate method on the binary activation map, which allows for the drawing of bounding boxes around each contour. The bounding boxes are then eliminated based on a size threshold, thus finalizing the detection process.

\section{Results}





\begin{figure} [t]
\centering
\begin{tabular}{cccc}
\includegraphics[width=0.3\textwidth]{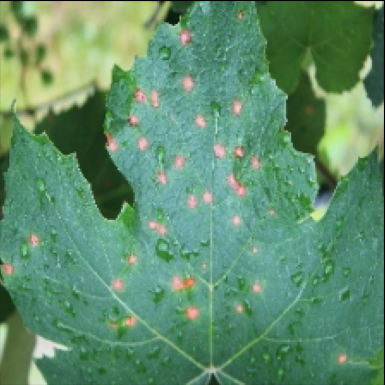} &
\includegraphics[width=0.3\textwidth]{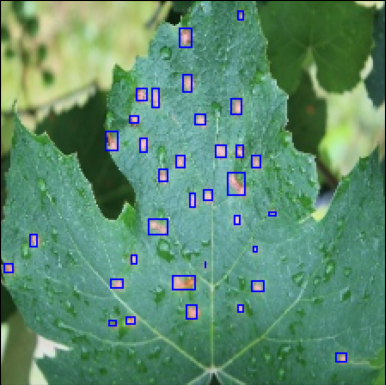} &
\includegraphics[width=0.3\textwidth]{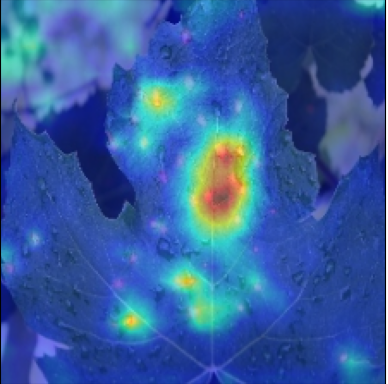} \\
\textbf{(a)}  & \textbf{(b)} & \textbf{(c)}  \\[6pt]
\end{tabular}
\begin{tabular}{cccc}
\includegraphics[width=0.3\textwidth]{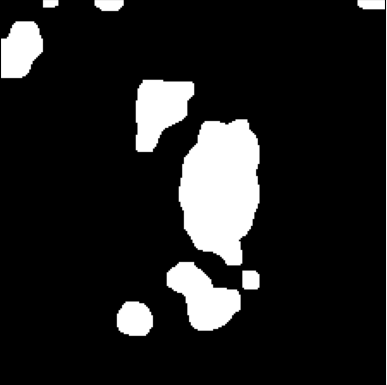} &
\includegraphics[width=0.3\textwidth]{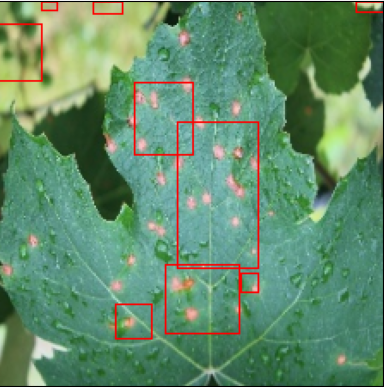} \\
\textbf{(d)}  & \textbf{(e)}  \\[6pt]
\end{tabular}
\caption{ \textbf{(a)} Diseased leaf image with lesions.
\textbf{(b)} Ground truth lesion bounding boxes.
\textbf{(c)} GradCAM output of the lesion classifier
\textbf{(d)} Segments obtained from the GradCAM output
\textbf{(e)} Predicted bounding boxes}
\label{fig:Name}
\end{figure}

The results of our experiments can be seen on Table \ref{tab:result}. The train dataset consists of 2100 images. We obtained an mAP score of 0.1967 with the IoU threshold being 0.001; it should be noted that GradCAM outputs have considerably low resolution, therefore it can be expected that the predicted bounding boxes are considerably larger than the ground truth boxes. Furthermore, these large boxes can cover multiple ground truth boxes and in such a situation, mAP tends to generate false negative predictions, even though the true predictions are included inside the large bounding box. For that, we also crafted a metric(which is referred to as the \textit{success rate}) where we count the ground truth boxes that are covered with a single large bounding box as true positive predictions. The success rate of capturing a lesion inside of a predicted bounding box is 0.4519. In the test dataset that consists of 400 images, we obtained mAP 0.1702 with 0.001 IoU. The success rate of capturing a lesion inside of a predicted bounding box is 0.4518.

\begin{table}[h]
\centering
\renewcommand{\arraystretch}{1.5}
\caption{Experimental Results on The Defined Metrics}
\begin{tabular}{c|c|c|}
\cline{2-3}
\textbf{}                                   & \textbf{Train Set (2100 image)} & \textbf{Test Set (400 image)} \\ \hline
\multicolumn{1}{|c|}{mAP (iou\_th=0.001)} & 0.1967                          & 0.1702                        \\ \hline
\multicolumn{1}{|c|}{Custom IoU Metric}     & 0.4519                          & 0.4518                        \\ \hline
\end{tabular}
\label{tab:result}

\end{table}

\section{Discussion}

In this study, we proposed a method for lesion detection on plant leaves using CAMs generated by a ResNet-18 classifier. By leveraging the capabilities of deep learning and CAMs, we aimed to accurately identify and localize lesions, providing a valuable tool for plant pathology and agricultural research.

While our study provides a promising approach for lesion detection on plant leaves, there are several avenues for future research and improvement. Firstly, the method can be further optimized and fine-tuned to achieve even higher accuracy and robustness. Exploring different CNN architectures and investigating the impact of hyperparameters can contribute to refining the lesion detection system. Additionally, expanding the dataset used for training the CNN classifier can enhance the model's ability to generalize across different plant species, diseases, and environmental conditions. In addition, layer selection for generating class activation maps is vital. We obtained the best results with intermediate layers. Lastly, research efforts should be directed towards the validation and evaluation of the proposed method on large-scale field trials and diverse plant species. Assessing its performance under different environmental conditions. 

In conclusion, our study presents a novel approach for lesion detection on plant leaves by utilizing CAMs generated by a ResNet-18 classifier. The significance of early and accurate lesion detection in plant pathology and agriculture cannot be overstated. The proposed method offers a valuable tool for researchers and farmers, enabling timely disease management and contributing to improved crop health and productivity. Future research can focus on optimizing the method, exploring additional data sources, and developing practical applications to further enhance its impact and usability.

\newpage 
\bibliographystyle{unsrtnat}
\bibliography{references}  






\end{document}